\documentclass[DIV=calc, paper=letter, fontsize=11pt]{scrartcl}	 

\usepackage[]{units}

\usepackage{lipsum} 
\usepackage[english]{babel} 
\usepackage[protrusion=true,expansion=true]{microtype} 
\usepackage{amsmath,amssymb,amsfonts,amsthm,bm,bbm} 

\usepackage[svgnames, table]{xcolor} 

\usepackage[hang, small,labelfont=bf,up,textfont=it,up]{caption} 
\usepackage{booktabs} 

\usepackage{fix-cm}
\usepackage[]{units}

\usepackage{subfig}
\usepackage{physics}

\usepackage{tabularx}

\usepackage[colorlinks=true, linkcolor=blue, citecolor=blue, filecolor=blue,
runcolor=blue, urlcolor = blue]{hyperref} 

\usepackage[capitalize, noabbrev]{cleveref}

\usepackage{fullpage}
\usepackage{sectsty} 

\usepackage{fancyhdr} 
\usepackage{lastpage}

\usepackage{nicefrac}

\usepackage{tikz}
\usetikzlibrary{calc,patterns,decorations.pathmorphing,decorations.markings}
\usepackage{pgfplots}

\usepackage{multicol}

\usepackage{float}

\usepackage[utf8]{inputenc}
\usepackage[english]{babel}
\usepackage{tikz}
\usepackage{comment}

\usetikzlibrary{arrows}

\usepackage{mathtools} 

\usepackage{algorithm, algpseudocode}

\usepackage[
backend=biber, 
style=numeric-comp,
sorting=none
]{biblatex}

\makeatletter
\newcommand{\multicolinterrupt}[1]{
\setcounter{tempcolnum}{\col@number}
\end{multicols}
#1
\begin{multicols}{\value{tempcolnum}}
}
\makeatother

\definecolor{issuePJA_color}{rgb}{1.0,0.0,0.0}

\definecolor{commentPJA_color}{rgb}{1.0,0.0,0.8}

\definecolor{issueQB_color}{rgb}{1.0,0.8,0.0}

\definecolor{commentQB_color}{rgb}{0.6,0.0,0.8}

\definecolor{rev_color}{rgb}{0.6,0.0,0.0}

\definecolor{atz_table1}{rgb}{0.85, 0.85, 0.85}
\definecolor{atz_table2}{rgb}{0.8, 0.8, 0.8}
\definecolor{atz_table3}{rgb}{0.75, 0.75, 0.75}

\newcommand{\mb}[1]{\mathbf{#1}}
\newcommand{\bsy}[1]{\boldsymbol{#1}}

\renewcommand{\vec}[1]{\mb{#1}}

\newcommand{\FirstForm}{\bsy{\mathrm{I}}}

\newcommand{\SecondForm}{\bsy{\mathrm{I\!I}}}

\newcommand{\R}{\mathbb{R}}

\newcommand{\pn}[1]{\left( #1 \right)}

\newcommand{\set}[1]{\left\{ #1 \right\}}

\newcommand{\mat}[1]{\ensuremath{ \begin{bmatrix} #1 \end{bmatrix} }}

\usepackage{graphicx}
\DeclareGraphicsExtensions{.png}

\usepackage{lettrine} 
\newcommand{\initial}[1]{ 
\lettrine[lines=3,lhang=0.3,nindent=0em]{
\color{DarkGoldenrod}
{\textsf{#1}}}{}}

\usepackage{caption}

\DeclareOldFontCommand{\bf}{\normalfont\bfseries}{\mathbf}

\usepackage{graphicx}

\usepackage{titling} 

\newcommand{\HorRule}{\color{DarkGoldenrod}

\rule{\linewidth}{1pt}} 

\pretitle{\vspace{-80pt} \begin{flushleft} \HorRule \fontsize{14}{14}
\color{DarkRed} \selectfont} 
\title{Geometric Neural Operators (GNPs) for Data-Driven \\ Deep
Learning of Non-Euclidean Operators}
\posttitle{\par\end{flushleft}} 

\preauthor{\vspace{-15pt} \begin{flushleft}\fontsize{12}{12} 
\color{DarkRed}} 
\author{B. Quackenbush$^{1}$, P. J. Atzberger$^{1,2}$ } 

\postauthor{\small \color{Black} 
\\ $[1]$ Department of Mathematics,
University of California Santa Barbara (UCSB). \\
$[2]$ Department of Mathematics, Department of Mechanical Engineering, 
University of California \\ Santa Barbara (UCSB); atzberg@gmail.com;
\url{http://atzberger.org/}
\vskip 1.5em
\vspace{-0.6cm}
\HorRule
\end{flushleft} 
}

\pgfplotsset{compat = 1.16}

\DeclareGraphicsExtensions{.png}

\begin{document}
\date{}
\maketitle

\thispagestyle{fancy}

\vspace{-1.75cm}

\initial{W}\textbf{e introduce Geometric Neural Operators (GNPs) for accounting
for geometric contributions in data-driven deep learning of operators.  We show
how GNPs can be used (i) to estimate geometric properties, such as the metric
and curvatures, (ii) to approximate Partial Differential Equations (PDEs) on
manifolds, (iii) learn solution maps for Laplace-Beltrami (LB) operators, and
(iv) to solve Bayesian inverse problems for identifying manifold shapes.  The
methods allow for handling geometries of general shape including point-cloud
representations.  The developed GNPs provide approaches for incorporating the
roles of geometry in data-driven learning of operators.
}

\setlength{\parindent}{5ex}

\section{Introduction}
Many data-driven modeling and inference tasks require learning operations on
functions~\cite{Bronstein2021,Izenman2012,Stuart2010,
Atzberger2022,Atzberger2020,Chen1995,Kovachki2021a}.  Problems involving
mappings between function spaces include learning solution operators for
Partial Differential Equations (PDEs) and Integral
Operators~\cite{Strauss2007,Pozrikidis1992,Audouze2009}, estimators for inverse
problems~\cite{Stuart2010,Kaipio2006,Gelman1995}, and data
assimilation~\cite{Stuart2010,Asch2016}.  For many of these tasks, there are
also significant geometric and topological
structures~\cite{Bronstein2021,Bronstein2017,Wasserman2018,Izenman2012}.
Sources of geometric contributions can arise both directly from the problem
formulation~\cite{Bronstein2021,Hackel2017,Atzberger2020,Lassila2010} or  from
more abstract
considerations~\cite{Wasserman2018,Fefferman2016,Audouze2009,Atzberger2022}.
For example, PDEs on manifolds or with domains having complicated
shapes~\cite{Atzberger2020,Nguyen2015}.  More abstract sources of geometric
structure also can arise, such as the subset of solutions of parametric PDEs
arising from smooth
parameterizations~\cite{Atzberger2022,Haasdonk2017,Quarteroni2015,
Lassila2013,Bhattacharya2021} or from qualitative analysis of dynamical
systems~\cite{Hirsch2012,Arnold2013}.  Related geometric problems also arise in
many inference settings, learning tasks, and numerical
methods~\cite{Bronstein2017,Wasserman2018,Meilua2023,Ortiz2016}.  This includes
approaches for handling point-clouds in shape classification~\cite{Hackel2017},
or in developing PDE solvers on manifolds~\cite{Atzberger2022}.  Deep neural
networks hold potential for significant impacts on these problems by providing
new approaches for non-linear approximations, learning  representations for
analytically unknown operations through training, providing accelerations of
frequent operations, or discovering geometric structures within
problems~\cite{Chui2018,Izenman2012,Meilua2023,Li2024,
Atzberger2020,Atzberger2022,Bronstein2021}.

We introduce a class of deep neural networks for learning operators leveraging
geometry referred to as \textbf{\textit{Geometric Neural Operators (GNPs)}}.
The GNPs introduce capabilities for incorporating geometric contributions as
features and as part of the operations performed on functions.  This allows for
handling functions and operations on arbitrary shaped domains, curved surfaces
and other manifolds.  This includes capturing non-linear and geometric
contributions arising in computational geometry tasks, geometric PDEs, and
shape reconstruction inverse problems.

Related work has been done on function operators and parameterized PDEs, but
primarily on euclidean
domains~\cite{Chen1995,Lu2021,Kovachki2023a,FNO_2020,
Li2020,OLearyRoseberry2022a}.
Many of the methods also are based on using linear 
approaches, such Proper Orthogonal Decomposition
(POD)~\cite{Chatterjee2000,Schmidt2020} or Dynamic Mode Decomposition
(DMD)~\cite{Schmid2010,Kutz2016,Schmid2022}.  More recently, neural networks
have been used for developing non-linear approximation approaches.  This
includes the early work in~\cite{Chen1995}, and more recent work on Neural
Operators~\cite{Lu2021,Kovachki2023a} and incorporating geometry in~\cite{Li2024}.  
A few more specialized realizations of
this approach are the Fourier Neural Operator
(FNO)~\cite{FNO_2020,Li2020,Li2021b,Li2024}, Deep-O-Nets~\cite{Lu2021}, and Graph
Neural Operators~\cite{GNO_2019}.  
GNPs handle the geometric contributions in addition to function inputs 
based on network architectures building on Neural 
Operators~\cite{Kovachki2023a,GNO_2019}.

We organize the paper as follows.  We discuss the formulation of Geometric
Neural Operators (GNPs) in Section~\ref{sec:geo_neural_op}.  We then develop
methods for training GNPs for learning geometric quantities from point cloud representations
of manifolds in Section~\ref{sec:geo_point_cloud}.  We show the GNPs can be
used to approximate Laplace-Beltrami (LB) operators and to learn solutions to
LB-Poisson PDEs in Section~\ref{sec:poisson_lb}.  We then show how GNPs can be
used to perform inference in Bayesian Inverse problems for learning manifold
shapes in Section~\ref{section:inverse}.  Our results show how GNPs can be used
for diverse learning tasks where significant contributions arise
in operations from the geometry.

\section{Geometric Neural Operators}
\label{sec:geo_neural_op}
For learning general non-linear mappings between infinite dimensional function
spaces on manifolds that incorporate geometric contributions, we build on the
neural operator framework~\cite{Kovachki2023a,Chen1995}.  In contrast to more
conventional neural
networks which map between finite dimensional vector spaces, we use approaches
that learn representations for operators that are not strictly tied to the underlying
discretizations used for the input and output functions.  

We consider geometric operators of the form $\mathcal{G}[w,\Phi] \rightarrow
u$.  This operator takes as input a function $w(\cdot)$, where 
$w: \R^{d_i} \rightarrow \R^{d_w}$ with $w \in \mathcal{W}$ for some function space
$\mathcal{W}$, and a geometric description $\Phi$ with $\Phi: \R^{d_i} \rightarrow \R^{d_s}$, $\Phi \in
\mathcal{S}$, and gives as output a function $u(\cdot)$ with $u: \R^{d_i}
\rightarrow \R^{d_u}$, $u \in \mathcal{U}$.  The operator can be expressed as a
mapping $\mathcal{G} : \mathcal{W}\times \mathcal{S} \rightarrow \mathcal{U}$.
This approach allows for
flexibility in learning dependencies on the geometry and allows for 
$\Phi$ to be formulated in a few different 
ways.  One approach is to provide information
by parameterizing the manifold geometry.  This can leverage one of the most common approaches 
using coordinate charts with $\Phi: \R^{d_i + 1}
\rightarrow \R^{d_g}$ where the extra component gives the chart index $I$.
In the case of a surface, we would have $d_i = 2$ and
$w(z_1,z_2,I)$ and $\Phi(z_1,z_2,I) \in \R^3$ with $\mb{z} \in \R^2$.  As another 
approach, the geometry
can be described by an embedding in $\R^{d_g}$ and the
collection of points $\mathcal{M} = \{\mb{x}\;|\; g_i(\mb{x}) = 0,\;
i=1,\ldots,k\}$.  In this case, $w = w(\mb{x})$ has inputs $\mb{x} \in
\mathcal{M}$ and we can use $\Phi(\mb{x}) = \mb{x}$.  As an alternative when 
geometric quantities and contributions are already known in advance, such as
an operator that only depends on the local principle curvatures 
$\kappa_1,\kappa_2$, we can simplify learning by 
letting $\Phi(\cdot) = [\kappa_1(\cdot),\kappa_2(\cdot)]$.  
This allows for a few different ways to explicitly incorporate geometric
contributions when learning operators. 
In the notation, we denote the combination of input function and geometric description 
by $a(\cdot) = [w(\cdot),\Phi(\cdot)]$ with $a:\;\R^{d_i} \rightarrow \R^{d_a}$, $a \in
\mathcal{A} = \mathcal{W}\times \mathcal{S}$.   

\begin{figure}[t]
\centering
\includegraphics[width=0.99\textwidth]{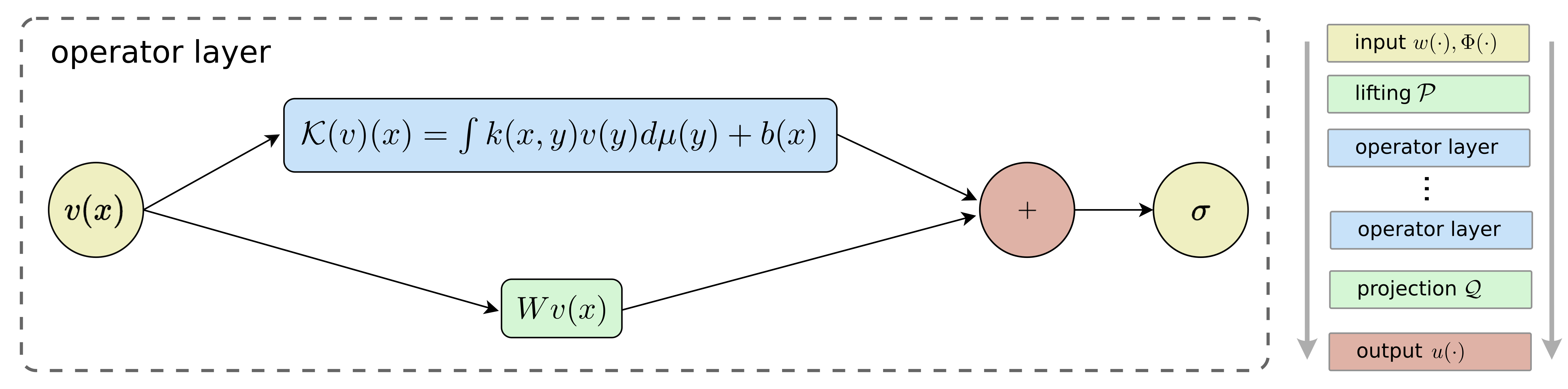}
\caption{\textbf{Deep Learning Methods for Operators.} An operator
layer is used as part of processing input functions.  For a function
$v(\cdot)$, an affine operation is performed based on 
integration against a kernel $k(x,y)$ and 
adding a bias $b(\cdot)$.  An additional skip connection with a local linear 
operator $W$ is also added to the pre-activation output of the layer.  
These linear operations are then followed by applying the activation 
function $\sigma(\cdot)$ \textit{(left)}.  In combination with the
operations of lifting $\mathcal{P}$ and projection $\mathcal{Q}$, these 
layers are stacked to process input functions to obtain deep learning methods for
approximating operators \textit{(right)}. 
}
\label{fig:op_layer}
\end{figure}

We approximate the geometric operators $\mathcal{G}: \mathcal{A} \to
\mathcal{U}$ by developing methods for learning a 
neural operator $\mathcal{G}_\theta: \mathcal{A} \to \mathcal{U}$
with parameters $\theta$.  
The $\mathcal{G}_\theta$ approximation consists of the following three 
learnable components (i)
performing a lifting procedure $\mathcal{P}$ for $a \in \R^{d_a}$ to a higher dimensional set of feature functions
$v_0 \in \R^{d_v}$ with $d_v \geq d_a$, (ii) performing compositions of layers consisting of  
a local linear operator $W$, integral operator $\mathcal{K}$, bias function $b(\cdot)$, and 
non-linear activation $\sigma(\cdot)$, to obtain
$v_{i+1} = \sigma \pn{Wv_i + \mathcal{K}[v_i] + b}$, 
and (iii) performing projection $\mathcal{Q}$ to a $\R^{d_u}$-valued function,
see Figure~\ref{fig:op_layer}.  The trainable components include the lifting procedure $\mathcal{P}$, 
kernel $k$, bias $b$ function, and local operator $W$ in the operator layers, and the projection 
$\mathcal{Q}$.  We collect all of these parameters into $\theta$. 
This gives a neural operator with $T$ layers of the general form 
\begin{equation}
\mathcal{G}_\theta^{(T)} = \mathcal{Q} \circ \sigma_T\pn{W_T + \mathcal{K}_T +
b_T} \circ \dots \circ \sigma_0\pn{W_0 + \mathcal{K}_0 + b_0} \circ
\mathcal{P}.
\end{equation}

The activation of the last layer $\sigma_T$ is typically taken to be the identity.
The special case of a neural operator with a single layer has the form 
$\mathcal{G}_\theta^{(1)} = \mathcal{Q} \circ \sigma\pn{W + \mathcal{K} + b}
\circ \mathcal{P}.$

For the linear operators $\mathcal{K}$, we consider primarily integral operators of the form
\begin{equation}\label{kernel_integral}
\mathcal{K}(v)(x) = \int_D k(x, y) v(y) \ d\mu(y).
\end{equation}
The $\mu$ is a measure on $D \subset \R^{d_{v}}$, $v:D \to \R^{d_{v}}$ is the
input function, and $k$ is a kernel $k(x, y) \in \R^{d_{v}} \times \R^{d_{v}}$.
For each layer $t$, we consider a trainable kernel $k = k(x,y;\theta_t)$
parameterized by $\theta_t$ for fully connected neural networks having
layer-widths $(d_a, n/4, n/2, n, d_v^2)$ for $n \in \mathbb{N}$.  

In practice, we fix $d_v$ to be the latent dimension of the nodal features
within the hidden layers and we use Lebesgue measure for $\mu$.  For activation
functions, we use the ReLU on all internal layers.  Other choices and
functional forms for $\mathcal{K}$ and the kernels $k$ 
also can be considered to further adapt techniques for 
special classes of problems.

\subsubsection{Approximating the Integral Operations}
For approximating the integral operations on general manifolds, we develop
methods for approximating the integral operations required to evaluate 
$\mathcal{K}$ building on graph neural operators~\cite{Kovachki2023a}.  
We further develop methods using sparse kernel evaluations 
and specialized constraints on the form of $k$ in the learned kernels.

As an initial approximation of the integral operator $\mathcal{K}$,
consider using $J$ sample points $\{x_k\}_{k=1}^J$ to obtain
\begin{equation}
\mathcal{\hat{K}}(v)(x_i) = \frac{1}{J} \sum_{j = 1}^J k(x_i, x_j) v(x_j),\,\,\, 
i = 1, \dots, J.
\end{equation}
Here, we assume the measure $\mu$ in $\mathcal{K}$ is normalized to have 
$\mu(D) = 1$.  Direct evaluation of these expressions gives a 
computational complexity $O(J^2)$.

\begin{figure}
\centering
\includegraphics[width=0.8\textwidth]{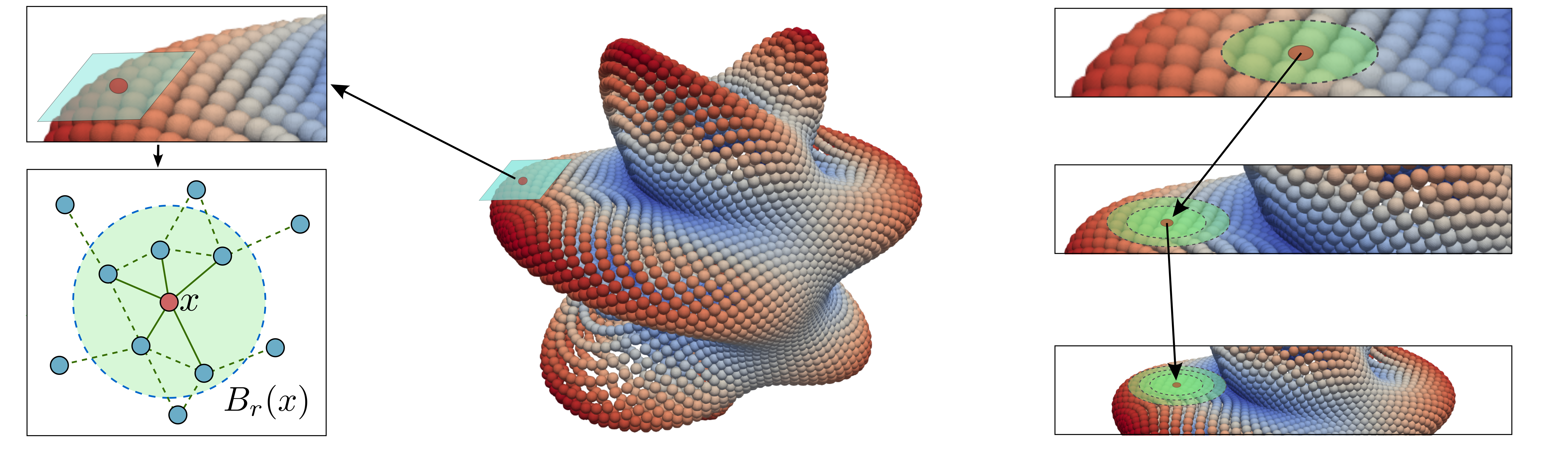}
\caption{\textbf{Approximating $\mathcal{K}[v]$: Graph-based Approaches and Truncations.} 
We develop methods for general manifolds, including with point-cloud
representations \textit{(middle)}.  We approximate integral operations by using graph neural
operators and message passing.  To evaluate
$\mathcal{K}[v](x)$ at node $x$, we use mean aggregation.  The nodes $x$ and $y$ 
have an edge connection in the graph when $\|x - y\| < r$.  The graph has
node attributes $v(y)$ and edge attributes $k(x,y)$.  We take the mean over 
all updates $k(x,y)v(y)$ for neighboring nodes $\mathcal{N}(x)$ to obtain 
the approximation of $\mathcal{K}[v]$.  To help make computations 
more efficient, we also truncate the neighborhood to a ball $B_r(x)$ of
radius $r$ \textit{(left)}.  In deep learning methods, stacking the 
operator layers increases successively in depth from the previous layers 
the effective domain of dependence of the overall operator \textit{(right)}. 
}
\label{fig:graph_approx}
\end{figure}

To help manage these computational costs we use a few approaches to control
the size of $J$.  This includes (i) using a sparse sub-sampling of the points 
$\{x_i\}$, and (ii) truncating the domain of integration to $S(x)$, such as 
a ball $B_r(x)$ of radius $r$, see Figure~\ref{fig:graph_approx}.  Using 
these approaches, we approximate the kernel operator 
$\mathcal{K}$ by 
\begin{equation}
\mathcal{\tilde{K}}[v](x) = \int_{S(x)} k(x, y) v(y) \ dy, \quad \forall x \in
D.
\end{equation}
The truncation $S(x) \subset D$ is given by $S:D \to \mathcal{B}(D)$ which maps
$x$ to the neighborhood $S(x)$.  The $\mathcal{B}(D)$ denotes the subsets of $D$ that
are Lebesgue measurable.  We denote the indicator function for such sets as $d\mu(x, y) =
\mathbbm{1}_{S(x)} dy$.  While more general choices can also be made,
we will primarily use $S(x) = B_r(x)$ for a ball of radius
$r$ centered at $x$.  We remark that while the truncations may seem to reduce
the domain of dependence of the operator to $B_r$, in fact, the overall
operator and information flow 
can still have longer-range dependencies.  In deep learning methods,
the operator layers are stacked which successively in the depth from
the previous layers increases the domain of dependence of the overall operator, see 
Figure~\ref{fig:graph_approx}.

We compute our approximations using random sampling and the following steps.
First, a graph is constructed using nodes $\set{x_1, \dots, x_D} \subset D$
with features $a(x_i)$ at $x_i$.  We construct directed edges to all nodes $y
\in S(x)$ with edge features $k(x,y)$.  For a given node $x_j$ having value
$v_t(x_j)$ and neighborhood $\mathcal{N}(x_j) = S(x_j) \cap \set{x_1, \dots,
x_J}$, we update the value at node $x_j$ using the mean aggregation 
\begin{equation}
\mathcal{\tilde{K}}[v_{t}](x_j) = \frac{1}{\abs{\mathcal{N}(x_j)}} \sum_{x_k \in
\mathcal{N}(x_j)}
k(x_j, x_k) v_t(x_k).
\end{equation}
We remark that for such integration operations on manifolds, the accuracy of
approximations depend on the sample points $\{x_i\}$, which must be taken
sufficiently dense to capture both the local shape of the manifold and the
local surface fields.  These properties can be characterized using quantities,
such as the fill-distance and estimated curvatures of the
manifolds~\cite{Atzberger2020}. 

\subsubsection{Kernel Restrictions with Factorizations and Block-Reductions}
\label{sec:kernel_factor}
The sampling of the kernel evaluations correspond to evaluating a linear
operator equivalent to the action of a $d_v \times d_v$ matrix.  In 
back-propagation, these calculations can readily exhaust memory 
during the gradient computations during training.  To help mitigate 
such computational issues as the latent dimension $d_v$ 
becoming large, we have developed further specialized functional forms
and restrictions for the trainable kernels.  

Since edges are more numerous than nodes, these contributions dominate
the calculations, and we seek restrictions that limit their growth. 
For this purpose, we consider kernels that can be factored as 
$k(x, y) = W_k \tilde k(x,y)$ where $\tilde k(x, y)$ is 
block diagonal 
\begin{equation}
\tilde k = \left[
\begin{array}{cccc}
B_1   &  0  & 0 & 0 \\
0   & B_2 & 0 & 0\\
0   &  0  & \ddots & 0 \\
0   &  0  & 0 & B_c 
\end{array}
\right].
\end{equation}
This consists of $c$ blocks denoted by $B_i = B_i(x, y)$ 
each having the shape $d_{v'} \times d_{v'}$. 
We remark that a block form for kernels was also considered and found to 
be helpful in the setting of
regular grids for a Fourier Neural Operator in~\cite{Guibas2021}.
A notable feature of our factorized form is that the trainable 
$W_k$ does not depend on the inputs $(x, y)$. 
The integral operator with this choice for $k$ can be expressed as 
\begin{equation}
\tilde{\mathcal{K}}_t[v_t] = \int_D k(x, y) v(y) \ dy 
= W_k \int_D \tilde k(x, y) v(y) \ dy
= \sum_{i=1}^c W_{k,i} \int_D B_i(x, y) v(y) \ dy.
\end{equation}
We have the weights $W_{k} \in \R^{d_v}\times\R^{cd_{v'}}$ and $W_{k,i} \in \R^{d_v}\times\R^{d_{v'}}$.
Since the kernels are represented by fully connected neural networks,
our factorized kernel form also provides further savings through the ability 
to use fewer total weights.  Even if considering only the final layer 
of a general kernel $k$, this would consist of
$w_n d_v^2$ parameters.  The $w_n$ is the width of the kernel network.
When we use $\tilde k$ instead, the final layer becomes $w_n c d_{v'}^2$
parameters, which given the quadratic dependence can be a significant 
savings.  Further, moving the matrix $W_k$ outside of the integral 
avoids having to apply $W$ directly to each edge weight $k(x,y)$. 
The factorization allows for only applying $W_k$ on the nodes of 
the graph after the aggregation step of the edge convolution, 
resulting in further savings.  These approaches taken together
are used in performing the steps in the operator layer to obtain
\begin{equation}
v_{t+1}(x) = \sigma_t \pn{W v_t(x) + \tilde{\mathcal{K}}_t[v_t] + b_t}.
\end{equation}
The $\tilde{\mathcal{K}}_t$ uses the block diagonal kernel $\tilde k_t$.
The factorized form for the kernel allows for use of larger latent-space 
dimensions and for deploying weights and computations into other parts of 
the neural operator approximation.  The approach also allows for overall
savings in memory and computational time during training.  

\section{Learning Geometric Quantities for Manifolds with Point-Cloud Representations}
\label{sec:geo_point_cloud}
Important contributions are made by geometry in 
many machine learning tasks, such as classifying shapes
or approximating solutions of PDEs on manifolds.
We develop geometric neural operators (GNPs) for estimating
geometric quantities, such as the metric and curvatures,
from point-cloud representations of manifolds.
To demonstrate the methods, we 
consider the setting of radial manifolds $\mathcal{M}$, and an
embedding
$\bsy{\sigma}(\theta, \phi)$ taking values from a coordinate chart $(\theta,\phi)$
into $\R^3$,~\cite{Atzberger2018}.  We focus on learning the first $\FirstForm$ and 
second $\SecondForm$ fundamental forms of differential geometry~\cite{Pressley2010}. 
These can be expressed in terms of the embedding map $\bsy{\sigma}$ as 
\begin{equation}
\FirstForm = \mat{E & F \\ F & G} = \mat{\bsy{\sigma}_\theta \cdot
\bsy\sigma_\theta & \bsy\sigma_\theta \cdot \bsy\sigma_\phi \\ \bsy\sigma_\phi
\cdot \bsy\sigma_\theta & \bsy\sigma_\phi \cdot \bsy\sigma_\phi},\hspace{1.25cm}
\SecondForm = \mat{L & M \\ M & N} = \mat{\bsy{\sigma}_{\theta \theta} \cdot
\bsy{n} & \bsy \sigma_{\theta \phi} \cdot \bsy n \\ \bsy \sigma_{\phi \theta}
\cdot \bsy n & \bsy \sigma_{\phi \phi} \cdot \bsy n}.
\end{equation}
The $\bsy{\sigma}_\theta = \partial_\theta \bsy{\sigma}$,
$\bsy{\sigma}_\phi = \partial_\phi \bsy{\sigma}$, and similarly for 
higher-order derivatives.  The 
outward normal $\bsy{n}$ is given by 
\begin{equation}
\bsy{n}(\theta, \phi) = \frac{\sigma_\theta(\theta, \phi) \times
\sigma_\phi(\theta, \phi)}{\norm{\sigma_\theta(\theta, \phi) \times
\sigma_\phi(\theta, \phi)}}.
\end{equation}
These forms can be used to construct the Weingarten map as $\bsy W
= \FirstForm^{-1} \SecondForm$.  We use $\bsy{W}$ to compute the 
Gaussian curvature 
\begin{equation}
K(\theta, \phi) = \text{det}(\bsy W(\theta, \phi)).
\end{equation}

\begin{figure}[h]
\centering
\includegraphics[width=0.99\linewidth]{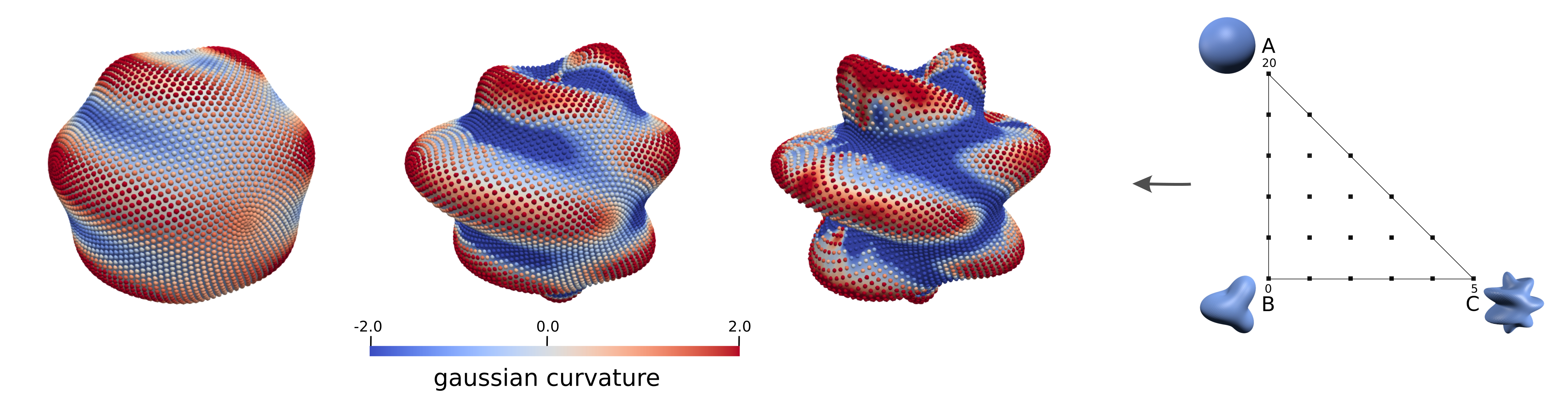}
\caption{\textbf{Geometric Quantities for Manifolds with Point-Cloud
Representations.}  We develop geometric neural operators (GNPs) to estimate
geometric quantities, such as the metric and curvatures, from manifolds with
point-cloud representations.  Shown are a few example shapes and their Gaussian
curvatures \textit{(left)}.  The methods are trained on random shapes obtained by
using barycentric coordinates to combine a collection of reference
manifolds $A,B,C$ depicted at the vertices with functional forms given 
in~\cite{Atzberger2018b} \textit{(right)}.
}
\label{manifold_gaussian}
\label{fig:barycentric_grid}
\end{figure}

We consider the learning tasks for a collection of different radial manifold shapes given by
combinations using barycentric coordinates arranged as in Figure~\ref{fig:barycentric_grid}.
Each of the radial manifolds are represented by a collection of spherical harmonic 
coefficients for the radial function $r(\theta,\phi)$, as in our prior work~\cite{Atzberger2018}.
These manifold shapes are sampled by using uniform random variables 
$u_1, u_2 \sim \mathcal{U}[0, 1)$, to obtain coefficients
$\bsy{d} = (1-\sqrt{u_1}) \bsy{a} + (1 - u_2)\sqrt{u_1}\bsy{b} + \sqrt{u_1}u_2
\bsy{c}$ where $\mb{a},\mb{b},\mb{c},\mb{d}$ are vectors for the collection of 
spherical harmonic coefficients. 

For our training dataset we sample $500$ manifold shapes.  We also 
sample $200$ manifold shapes as a test dataset.  
We should mention these intrinsic manifold shapes from the spherical harmonics will
primarily be used in our analysis and are treated distinct from the sampled 
point-cloud representations of the shapes used in training and testing. 

To obtain point-cloud representations for each of the manifold shapes 
we use the vector of spherical harmonic coefficients $\bsy{d}$ 
and a separate random sampling of $N=1024$ points $\{\mb{x}_i\}_{i=1}^N$ 
on the sphere with $\mb{x}_i$ projected to the manifold surface.  
We then use the spherical harmonic interpolation to obtain training 
data for the geometric quantities $\mb{z} = \mb{z}(\mb{x}_i;\mb{d}) = (x, y,z,E, F, G, L, M, N, K)$
for the manifold surface with shape $\mb{d}$ and point-cloud sample at location
$\mb{x}_i = (x,y,z)$, and $E,F,G,L,M,N,K$ from the first and second fundamental 
forms $\FirstForm,\SecondForm$.

We also add noise to the datasets to obtain training samples
$\mb{\tilde{z}} = \mb{z} + \bsy{\xi} = (x, y, z, E, F, G, L, M, N, K) + \bsy{\xi}$, where
$\bsy{\xi} \sim \mathcal{N}(0, \epsilon^2\Lambda)$ is Gaussian noise. 
The $\epsilon$ gives the relative noise for the reference scales $\Lambda$ given by
$\Lambda = \text{diag}(\{\lambda_i\})$ with 
$\lambda_i = \norm{(x, y, z)}_2^2,\;  i = 1, 2, 3;\, \lambda_i = \norm{(E, F,
G)}_2^2,\;  i = 4, 5, 6; \, \lambda_i = \norm{(L, M, N)}_2^2,\;  i = 7, 8, 9; \,
\lambda_i = \abs{K}^2,\; i = 10. $  To further test robustness of the methods,
we also consider two additional datasets with (i) uniform 
noise at 1\% added to all data components for each manifold shape, and (ii) by creating 
noisy perturbed outlier points by adding noise at 10\% to $50$ randomly chosen data 
points in each manifold shape.

We train the geometric neural operators (GNPs) to learn the geometric
quantities from the point cloud representation $\set{\bsy{x}_j}_{j=1}^N$ 
with $N=1024$ surface points for a manifold $\mathcal{M}^{(i)}$. 
The GNP $u=\mathcal{G}_\theta[\{\mb{x}_j\}]$ must learn from the point cloud $\{\mb{x}_j\}$
the fundamental forms $\FirstForm,\SecondForm$ by learning a mapping $u$
\begin{equation}\label{form_vector}
\bsy{x}_j \mapsto (E(\bsy x_j), F(\bsy x_j), G(\bsy x_j), L(\bsy x_j), M(\bsy
x_j), N( \bsy x_j), K( \bsy x_j)).
\end{equation}
We also include in the training learning the Gaussian curvature $K$.
We use a few different architectures of our GNPs with $d_v=64$, kernel width up to
$w_n=256$, and depth up to $d_n =10$ integral operator layers.
We use for our loss function the $L^2$-norm for comparing the GNP predicted 
values of the model $\mathcal{G}_\theta$ with the true values from the 
shapes obtained from the spherical harmonics representations.  We trained using optimization methods 
based on stochastic gradient descent with momentum using the 
Adam method~\cite{Kingma2014}.

\begin{table}[h]
\centering
{\fontsize{10.0pt}{12.0pt}\selectfont
\begin{tabular}{| l l | l  l | l  l |}
\hline
\rowcolor{atz_table3}
\textbf{depth} & \textbf{neurons} & \textbf{training error} & \textbf{test error} & \textbf{training error} & \textbf{test error} \\
\hline
\rowcolor{atz_table1}
\multicolumn{2}{|l|}{\textbf{architecture}} 
&\textbf{full kernel} &&\multicolumn{2}{|l|}{\textbf{factorized block-kernel}}\\
10&256& 2.52e-02$\pm$6.5e-04 & \textbf{3.82e-02$\pm$1.7e-03} & 4.23e-02$\pm$7.6e-04 & \textbf{5.19e-02$\pm$9.3e-04} \\
10&128& 4.36e-02$\pm$6.0e-04 & 5.57e-02$\pm$5.9e-04 & 7.73e-02$\pm$3.9e-03 & 8.54e-02$\pm$3.1e-03 \\
8&256& 2.83e-02$\pm$2.3e-04 & 4.13e-02$\pm$1.3e-03 & 5.08e-02$\pm$1.0e-03 & 6.00e-02$\pm$9.9e-04 \\
8&128& 5.08e-02$\pm$1.3e-03 & 6.29e-02$\pm$1.7e-03 & 8.94e-02$\pm$2.3e-03 & 9.59e-02$\pm$1.8e-03 \\
\rowcolor{atz_table1}
\multicolumn{6}{|l|}{
\textbf{1\% noise for all points}}\\
10&256& 6.31e-02$\pm$6.3e-04 & \textbf{8.78e-02$\pm$1.6e-03} & 8.11e-02$\pm$1.1e-03 & \textbf{9.55e-02$\pm$9.8e-04} \\
10&128& 8.00e-02$\pm$8.1e-04 & 9.83e-02$\pm$7.0e-04 & 1.11e-01$\pm$2.5e-03 & 1.20e-01$\pm$1.2e-03 \\
8&256& 6.68e-02$\pm$9.8e-04 & 8.97e-02$\pm$7.1e-04 & 8.63e-02$\pm$9.7e-04 & 9.79e-02$\pm$8.3e-04 \\
8&128& 8.75e-02$\pm$5.7e-04 & 1.03e-01$\pm$1.4e-03 & 1.18e-01$\pm$3.1e-03 & 1.26e-01$\pm$3.8e-03 \\
\rowcolor{atz_table1}
\multicolumn{6}{|l|}{
\textbf{10\% noise for 5\% of points}}\\
10&256& 1.00e-01$\pm$6.2e-04 & 1.49e-01$\pm$1.9e-03 & 1.26e-01$\pm$6.4e-04 & \textbf{1.49e-01$\pm$2.5e-03} \\
10&128& 1.22e-01$\pm$1.6e-03 & 1.54e-01$\pm$2.5e-03 & 1.49e-01$\pm$1.1e-03 & 1.67e-01$\pm$2.5e-03 \\
8&256& 1.06e-01$\pm$5.1e-04 & \textbf{1.48e-01$\pm$7.2e-04} & 1.31e-01$\pm$1.3e-03 & 1.52e-01$\pm$2.7e-03 \\
8&128& 1.28e-01$\pm$1.6e-03 & 1.59e-01$\pm$3.6e-03 & 1.58e-01$\pm$1.4e-03 & 1.76e-01$\pm$3.0e-03 \\
\hline
\end{tabular}
}
\caption{\textbf{Results for Learning Joint Geometric Properties.} 
The GNPs were trained with full and factorized kernels to estimate metrics and curvatures from manifolds 
with point-cloud representations.  The methods were tested on random shapes obtained by
sampling from shapes in Figure~\ref{fig:barycentric_grid}.
Results are reported for cases with and without noise $\epsilon$,
with the most accurate result in bold for each study.  We report the mean  
$\pm$ the standard deviation of the results over $5$ training runs.   The sampling resolution for
each shape was $N=1024$ and a radius of $r=0.4$ for the kernel integrations. 
The GNPs were trained for $500$ epochs with an initial learning rate of 
$10^{-4}$ which was halved every $100$ epochs.  ReLU activations 
were used both in the kernel networks and in the operator layers.  
} 
\label{table:geometric_quant}
\end{table}

We show results for training the GNPs 
with different choices for the neural network 
architectures and other hyper-parameters in Table~\ref{table:geometric_quant}.
We varied in the studies the 
width $w_n$ of the neural networks between $128$ and $256$.  We also varied the 
depth $T$ of the network from $8$ to $10$ layers which from stacking 
increases the effective range of the domain of dependence of the overall operator, 
see Figure~\ref{fig:graph_approx}. 
The results show the GNPs can learn from the point-cloud 
accurate representations simultaneously the metric 
components $E,F,G$ of the first fundamental form $\FirstForm$  and the
curvatures components $L,M,N,K$ of the second fundamental form $\SecondForm$
and Gaussian curvature. 
This provides the basis for performing further many downstream
tasks and analysis using concepts from differential geometry 
and the trained GNPs. 
The trained GNPs had an overall $L^2$-error around $5.19 \times 10^{-2}$ when there 
were no noise perturbations.  In the case of $1\%$ noise perturbations in 
the training dataset, we find an $L^2$-error of around $9.55 \times 10^{-2}$.  
We find in the case of outliers the $L^2$-error becomes around 
$1.49 \times 10^{-1}$. While there is some degradation in accuracy when 
noise is added to the dataset, the results show the methods are overall
robust providing decent estimates of the geometric quantities.    
We also find our factorized block-kernel methods can train to a comparable level
of accuracy as the full kernel and 
run about three times faster during training.

The results 
indicate the GNPs can be trained to obtain robust methods for 
estimating the metrics and curvatures of manifolds from their
point-cloud representations.

\section{Learning Solution Maps of Partial Differential Equations (PDEs) on
Manifolds: Laplace-Beltrami-Poisson Problems}
\label{sec:poisson_lb}

We develop GNPs to learn the solution map for Laplace-Beltrami (LB) PDEs on manifolds.  We consider
Laplace-Beltrami-Poisson (LB-P) problems of the form
\begin{equation} \label{equ:pde_lb}
\left\{\begin{array}{lll}
\Delta_{LB}\; u & = & -f \\
\int_\mathcal{M} u(\bsy {x}) \ d \bsy{x} & = & 0
\end{array}
\right\}.
\end{equation}
The $\mathcal{M}$ denotes the manifold, $\Delta_{LB}$ the Laplace-Beltrami
operator, $u$ the solution, and $f$ the right-hand-side (rhs) data.  When
considering closed manifolds $\mathcal{M}$, there can be additional constants
required in geometric PDEs related to the topology~\cite{Atzberger2018b,Atzberger2018}.  
For the 
spherical topology here, there is one additional constant which we determine by
imposing that the solution integrates to zero. 

We aim to learn the solution maps $u = -\Delta_{LB}^{-1} f$ for the PDE on the 
manifold surface  
for each of the reference shapes 
$A, B, C$ shown in \cref{fig:barycentric_grid}.  We will train separate GNPs 
for each manifold to effectively learn a solution operator $u = \mathcal{G}_{\theta}[\{\mb{x}_j\},-f]$.  
For this geometric elliptic PDE, such an operator is related to the 
Green's functions on the manifold.  For this purpose, we will train by generating
input functions using random samplings of $f$ of the form  
\begin{equation} 
f(\bsy{x} ; \bsy {\bar x},\sigma) = \frac{1}{\sqrt{2\pi \sigma^2}} \exp\pn{-\frac{1}{2
\sigma^2} \norm{\bsy x - \bsy{\bar x}}^2}  - c_0(\sigma).
\label{equ:f_rhs_sample}
\end{equation}
The $f$ have parameters $\sigma = 0.15$ and $\bsy{\bar x}$.  For the 
training data, we interpolate $f$ using 
spherical harmonics with $2030$ Lebedev nodes and 
hyper-interpolation~\cite{Atzberger2018,Atzberger2018b}.
To obtain random samples for $f$, we sample Gaussians 
$\bsy y \sim \mathcal{N}(0, I)$ that are radially projected onto the manifold 
$\mathcal{M}$ to obtain samples for $\bsy{\bar x}$.  We further choose $c_0(\sigma)$ 
so that the spherical harmonic interpolations of $f$ always have a zero spatial average
$ \frac{1}{|\mathcal{M}|}\int_{\mathcal{M}} f(\mb{x}) dx = 0$.
To obtain the input-solution pairs $(f,u)$ for the training data, we solve
the Laplace-Beltrami equations~\ref{equ:pde_lb}. This is done numerically
by building on our prior work on 
spectral methods for spheres based on Galerkin truncations over 
the spherical harmonics with $N=2023$ Lebedev nodes~\cite{Atzberger2018b,Lebedev1976}.  
\begin{figure}[h!]
\centering
\includegraphics[width=0.99\linewidth]{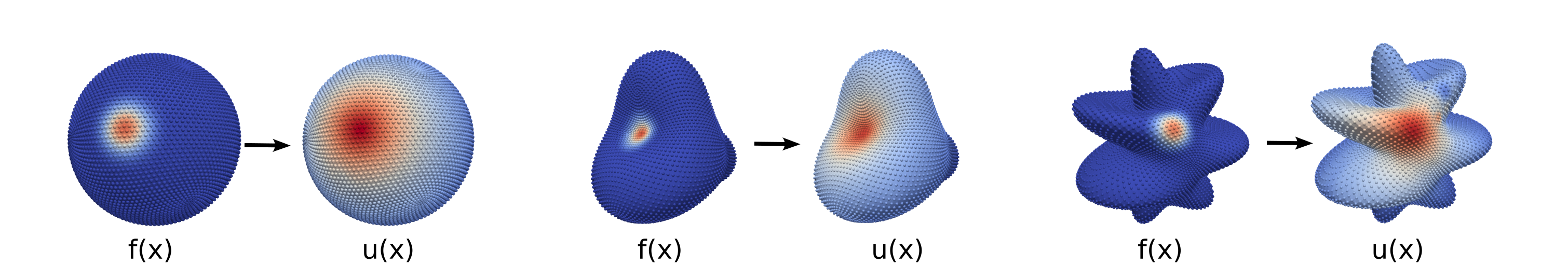}
\caption{\textbf{Training Geometric Operators for Solving PDEs on Manifolds.}  We train
neural geometric operators (GNPs) for the solution map of the 
Laplace-Beltrami PDE.  For the reference manifolds $A,B,C$ 
we show some example right-hand-sides (rhs) $f(x)$ and the corresponding solutions 
$u(x)$.  We train with locations of the Gaussian for the rhs varied over the surface
and over the collection of shapes.   }
\label{manifold_gaussian}
\end{figure}
We remark that one could in principle start with $u$ and compute the 
action of the differential operator  $-\Delta_{LB}[u]$ to obtain $f$.  However, in practice, 
we find from the differentiations involved that this is much more sensitive to noise
resulting in poorer quality training sets.  This indicates
when performing operator training, it is preferred to learn by 
sampling the underlying approximation of the solution map when 
it is available.  

To train and test our GNPs, we use
$1000$ random samples of $f$ to obtain the training set 
and another $200$ samples for testing.  We consider the final 
resolutions for the input-solution pairs $(f,u)$ using 
point-clouds with $N=1024$ spatial sample locations. 
Our loss function are based on the $L^2$-norm 
integrated over the manifold surface for the difference $u - \tilde{u}$ between the GNP predicted 
solution $\tilde{u}$ and the spectral solution of the PDE $u$.  We trained using optimization methods 
based on stochastic gradient descent with momentum using the Adam method~\cite{Kingma2014}.  We used 
$300$ epochs with a batch size of $1$ along with an initial
learning rate of $10^{-4}$ that was halved every $75$ epochs.

We show results of our training of GNPs for each of the manifolds $\mathcal{M}$
in Table~\ref{table:LB_A_full}.  We considered GNPs with a few different choices
for the neural network architectures.  We find our factorized block-kernel methods
perform comparable to the full kernel GNPs.  The accuracy over all the manifolds 
is found to be $9.03 \times 10^{-2}$.  The most accurate results of 
$1.07 \times 10^{-2}$ were obtained for Manifold $A$.  In this case, the manifold is
a sphere and the GNPs appear to benefit from the symmetry.   We further see in this case 
the depth and width of the neural networks do not have as much impact on the accuracy.  
As the shapes become more complicated, such as for Manifold $B$ and Manifold $C$,
the learning problem becomes more challenging.  This would be expected given the more 
heterogeneous contributions of the curvature within the differential operators in 
the Laplace-Beltrami 
equation~\ref{equ:pde_lb}.  
Relative to Manifold $A$, the Manifold $B$ is less symmetric and   
we find an accuracy of $3.75 \times 10^{-2}$.      
We see the depth and width of the GNPs has a more significant impact on the accuracy,
especially for the factorized case.  This could in part arise from the role 
depth plays in increasing the domain of dependence of
the operator.   For geometric quantities and computing differential operations,
there could be benefits from having a wider range of points over which to make estimates.  
In the case of the more complicated Manifold $C$, we find an 
accuracy of $9.03 \times 10^{-2}$.  We see the depth and width of the 
neural network again has a significant impact on the accuracy.  Overall, the results
indicate that GNPs can learn solution maps over a wide variety of shapes 
for geometric PDEs.

\begin{table}[h!]
\begin{center}
{\fontsize{10.0pt}{12.5pt}\selectfont
\begin{tabular}{| l  l | l  l | l  l |}
\hline
\rowcolor{atz_table3}
\textbf{neurons} & \textbf{depth} & \textbf{training error} & \textbf{test error} & \textbf{training error} & \textbf{test error} \\
\hline
\rowcolor{atz_table1}
\multicolumn{2}{|l}{\textbf{manifold A}} &\multicolumn{2}{|l}{\textbf{full kernel}}&\multicolumn{2}{|l|}{\textbf{factorized block-kernel}} \\
256&10& 7.50e-03$\pm$7.1e-04 & \textbf{8.98e-03$\pm$8.5e-04} & 9.72e-03$\pm$4.5e-04 & \textbf{1.07e-02$\pm$3.9e-04 }\\
128&10& 8.25e-03$\pm$5.8e-04 & 9.66e-03$\pm$6.5e-04 & 1.29e-02$\pm$3.6e-04 & 1.37e-02$\pm$4.3e-04 \\
256&8& 9.42e-03$\pm$7.0e-04 & 1.09e-02$\pm$7.5e-04 & 1.13e-02$\pm$8.6e-04 & 1.23e-02$\pm$9.3e-04 \\
128&8& 1.03e-02$\pm$4.6e-04 & 1.16e-02$\pm$4.7e-04 & 1.50e-02$\pm$7.4e-04 & 1.59e-02$\pm$7.7e-04 \\
\rowcolor{atz_table1}

\multicolumn{6}{|l|}{\textbf{manifold B}} \\
256&10& 1.85e-02$\pm$1.1e-03 & \textbf{3.32e-02$\pm$6.9e-04} & 2.74e-02$\pm$1.3e-03 & \textbf{3.75e-02$\pm$1.4e-03} \\
128&10& 2.45e-02$\pm$1.2e-03 & 3.74e-02$\pm$1.4e-03 & 3.75e-02$\pm$1.0e-03 & 4.40e-02$\pm$9.1e-04 \\
256&8& 2.43e-02$\pm$9.7e-04 & 3.84e-02$\pm$7.9e-04 & 3.41e-02$\pm$1.0e-03 & 4.24e-02$\pm$1.0e-03 \\
128&8& 3.07e-02$\pm$4.5e-04 & 4.19e-02$\pm$6.0e-04 & 4.40e-02$\pm$8.7e-04 & 4.93e-02$\pm$8.8e-04 \\
\rowcolor{atz_table1}

\multicolumn{6}{|l|}{\textbf{manifold C}} \\
256&10& 3.93e-02$\pm$1.9e-03 & \textbf{8.31e-02$\pm$2.1e-03} & 6.48e-02$\pm$2.8e-03 & \textbf{9.03e-02$\pm$1.1e-03} \\
128&10& 5.31e-02$\pm$3.5e-03 & 8.61e-02$\pm$2.2e-03 & 8.58e-02$\pm$7.8e-04 & 1.03e-01$\pm$1.3e-03 \\
256&8& 5.61e-02$\pm$1.5e-03 & 9.14e-02$\pm$1.3e-03 & 8.02e-02$\pm$1.5e-03 & 1.00e-01$\pm$1.4e-03 \\
128&8& 6.84e-02$\pm$2.6e-03 & 9.39e-02$\pm$2.0e-03 & 9.97e-02$\pm$1.3e-03 & 1.14e-01$\pm$1.7e-03 \\
\hline
\end{tabular}
}
\end{center}
\caption{
\textbf{Results for Learning Laplace-Beltrami Solution Operator.}
For the manifolds $A,B,C$, we show training and test errors of 
GNPs for the learned solution operator for the Laplace-Beltrami 
PDE in equation~\ref{equ:pde_lb}.  We report the mean  
$\pm$ the standard deviation of the results over $5$ training runs.
We show a few different choices 
for the GNP architectures with the most accurate results shown 
in bold for each study.   
}
\label{table:LB_A_full}
\end{table}

\section{Estimating Manifold Shapes: Bayesian Inverse Problems using
Observations of Laplace-Beltrami Responses} 
\label{section:inverse}

We developed GNPs for estimating manifold shapes from observations of the
action of the Laplace-Beltrami operator on the manifold.  We use as
observations the input-solution pairs $(u,f)$ for the Laplace-Beltrami-Poisson
PDE in equation~\ref{equ:pde_lb}.  For this purpose, we train GNPs to be used
in conjunction with solving a Bayesian Inverse Problem.  This consists of the
following steps
(i) formulate a prior probability distribution over the shape space, 
(ii) observe the Laplace-Beltrami responses for a collection of samples $\{(u^{(i)},f^{(i)})\}_{i=1}^m$,
(iii) use Bayes' rule to obtain a posterior probability distribution, 
(iv) perform optimization to find the most likely manifold shape under the posterior distribution. 
We train GNPs over both manifold shapes and Laplace-Beltrami responses.
The GNPs serve to map manifold shapes to Laplace-Beltrami responses to compute likelihoods.
As we shall discuss in more detail below,  
the likelihoods $p(\mathcal{M}|(u,f))$ are taken as Gaussians based on the $L^2$-norm for the difference
$\tilde{u} - u$ between the GNP predicted solution $\tilde{u} = \mathcal{G}_{\theta}[\{\mb{x}_j^\mathcal{M}\},f]$ and the 
data $u$.  We use this in conjunction with Bayesian inference to assign a posterior distribution
and obtain a Maximum A Posteriori (MAP) estimate for the 
shape $\mathcal{M}$.  
We can also use the GNPs to obtain an estimate of the full posterior distribution by using them 
for Monte-Carlo sampling.  

We consider the inverse problem of using observation data $\mathcal{D}$ to 
recover $\mathcal{M}$. 
We consider radial manifolds $\mathcal{M}$ and the responses of 
Laplace-Beltrami operators $\Delta_{LB}[u; \mathcal{M}]$ when applied to 
functions $u:\mathcal{M} \to \R$.
The observation data is $\mathcal{D} = \{\left(u^i(\mb{x}_j),
f^i(\mb{x}_j)\right)\}_{i,j}$ for $j=1,\ldots,N$ with $\{\mb{x}_j\}$ 
sampled uniformly on $S^2$ and radially projected
to $\mathcal{M}$ and where $\Delta_{LB}[u^i;\mathcal{M}] = -f^i$.
 The manifold shapes use the barycentric coordinates 
in \cref{fig:barycentric_grid}. As discussed in
Section~\ref{sec:geo_point_cloud}, the manifold shapes can be described by
spherical harmonic coefficient vectors $\vec{d}$.  We obtain responses by considering 
rhs-functions $f$ as in \cref{equ:f_rhs_sample}.
To obtain the response training data pairs $\{(u^i,f^i)\}$, 
we use the spectral solvers in our prior work~\cite{Atzberger2018,Atzberger2018b}. 
We obtain $f^i$ by sampling $\mb{\bar{x}}^i$ in equation~\ref{equ:f_rhs_sample} at   
$M=194$ Lebedev nodes for the order $12$ hyper-interpolation~\cite{Atzberger2018b}. 
Our spectral solver then yields a spherical harmonics representation of
 $u^i$.  We further sample $21$ manifold shapes from the barycentric 
coordinates as in \cref{fig:barycentric_grid}.  This is used to construct the full set of 
Laplace-Beltrami response pairs $\{(u^i,f^i)\}$ across all shapes.  This yields a training
set consisting of $4074$ samples of Laplace-Beltrami response pairs.

We train our GNPs by using data samples of the from  $(\mathcal{M}^{(i)}, u^{(i)}, f^{(i)})$.  
This consists of  a 
point-cloud sampling of the manifold geometry $\mathcal{M}^{(i)} = \{\mb{x}_j\}_{j=1}^N$
and a Laplace-Beltrami response pair $(u^{(i)},f^{(i)})$. 
We sample the geometry and responses using a collection of $N=1024$ points $\bsy{x}^{(i)}_j
\in \mathcal{M}^{(i)}$ to yield
$\left(\bsy{x}^{(i)}_j,
u^{(i)}(\bsy{x}^{(i)}_j), f^{(i)}(\bsy{x}^{(i)}_j)\right)$.  The GNPs are trained to learn 
the solution operator $\left(\bsy{x}^{(i)}_j, f^{(i)}(\bsy{x}^{(i)}_j)\right) \mapsto
u^{(i)}\left(\bsy{x}^{(i)}_j\right).$  We use as our loss function the $L^2$-norm of
the difference $\tilde{u} - u$ between the GNP prediction of the solution $\tilde{u}$
and the solution $u$ obtained from the spectral solvers.
For our GNP architectures we used $d_v=64$, kernel widths
$w_n=256$, and depths of $d_n=10$ for the integral operator layers.  For the optimization we use 
stochastic gradient descent with momentum based on Adam~\cite{Kingma2014}. 
We trained using $300$ epochs with 
a batch size of $1$.  Our learning rate was set to $10^{-4}$ and 
was decreased to $10^{-6}$ over $60$ epochs using cosine annealing~\cite{Loshchilov2016}.
In our cosine annealing, we cycled the learning rate every $60$ epochs to be restarted back 
at $10^{-4}$. 
 
\begin{figure}
\centering \includegraphics[width=0.7\textwidth]{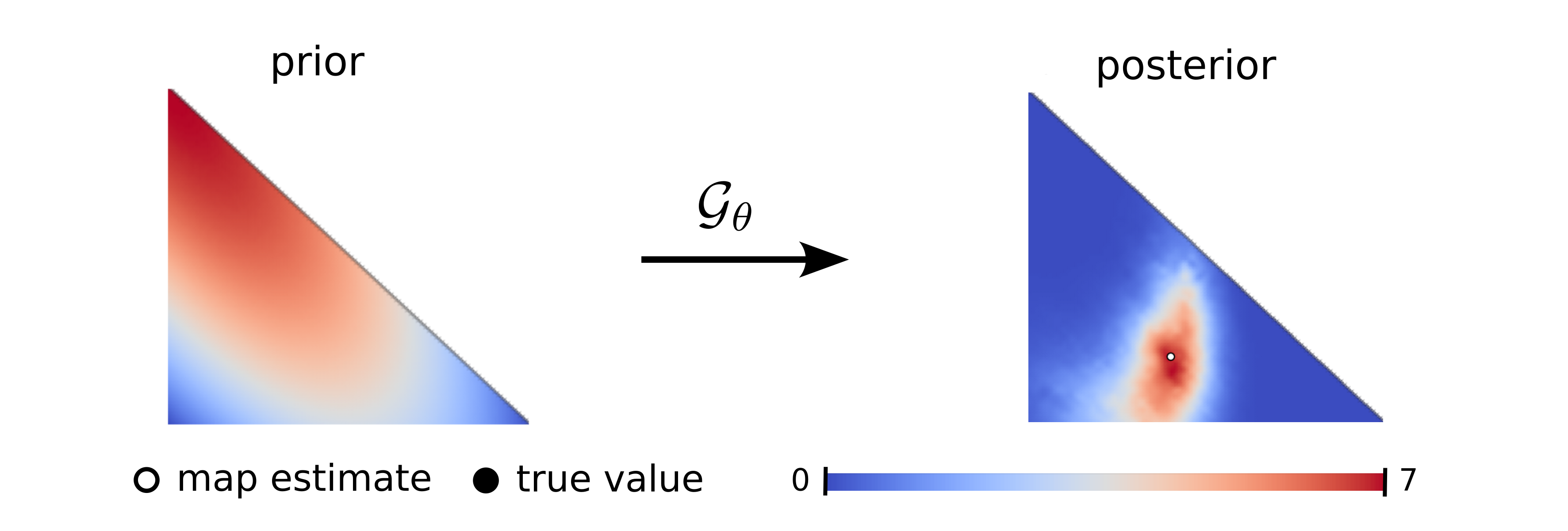}
\caption{\textbf{Manifold Shape Estimation: Prior Distribution.}
The prior distribution over manifolds $\mathcal{M}$ based on the radial
shape covariances when $\beta = 1$.  This is used in combination
with Bayes' Rule and the geometric neural operators (GNPs) to obtain
a posterior distribution over manifold shapes.  } \label{fig:prior}
\end{figure}

We used our trained GNPs $\mathcal{G}_{\theta}$ to perform 
Bayesian inference.  This requires developing a prior distribution $p(\mathcal{M})$
and likelihood distribution $p(\mathcal{D}|\mathcal{M})$.  We then seek to estimate 
a posterior distribution $p(\mathcal{M} | \mathcal{D}) $ from the observation
data $\mathcal{D}$ to assign a probability that the manifold shape 
was $\mathcal{M}$.  We use Bayes' Rule to obtain
\begin{equation}
p(\mathcal{M} | \mathcal{D}) = \frac{1}{Z} p(\mathcal{D} | \mathcal{M}) p(\mathcal{M}),
\end{equation}
where $Z$ is a normalization factor so the probabilities total to one.

We solve the inverse problem in practice by using our GNPs $\mathcal{G}_\theta$
as surrogate models for the Laplace-Beltrami responses.  We obtain likelihoods
by considering  
\begin{equation}
\xi^2 = \frac{1}{M} \sum_{i = 1}^M \frac{\norm{u^{(i)} -
\mathcal{G}_\theta[\{\mb{x}_j^{(i)}\},f^{(i)}]}^2}{\norm{u^{(i)}}^2}.
\end{equation}
The sum here is taken over the $M$ samples of the responses.
We approximate the $L^2$-norms by performing further summation over the point-cloud samples at 
$\{\mb{x}_j\}$.  This yields the likelihood 
\begin{equation}
p \pn{ \mathcal{D} | \mathcal{M}} = \pn{2 \pi \sigma^2}^{-1/2} \exp
\pn{-\frac{\xi^2}{2\sigma^2}}.
\end{equation}
We take here  $\sigma^2 = 10^{-3}$.

We develop a prior distribution for use over the manifolds $\mathcal{M}$ 
of the form
\begin{equation} \label{prior_eq}
p(\mathcal{M}) \propto \exp \pn{-\beta F(\mathcal{M})}.
\end{equation}
We choose $F$ to characterize the complexity of the manifold geometry by 
using the radial shape functions $r_\mathcal{M}(\theta, \phi)$.  We use
the covariance of the radial function $r_\mathcal{M}$ to obtain 
$$F(\mathcal{M}) = \int \left(r_\mathcal{M}(\theta,\phi) - \mu_r\right)^2
d\theta d\phi. $$  The mean radius is given by 
$\mu_r = \int r_\mathcal{M}(\theta,\phi) d\theta d\phi $.
The $F$ can be thought of as a free energy for the manifold shapes. 
The $\beta$ acts like an inverse temperature
that controls the characteristic scales at which to put emphasis 
on the differences in the radial covariances.  We use as our default value 
$\beta = 1$.  This provides a prior distribution that serves to regularize results
toward simpler shapes, such as a sphere which has the smallest radial covariance.
We show our prior distribution over the barycentric interpolated shape space in 
Figure~\ref{fig:prior}.

\begin{table}[h]
\begin{center}
{\fontsize{9.0pt}{11.0pt}\selectfont
\begin{tabular}{|l|llll|llll|}
\hline
\rowcolor{atz_table3}
\textbf{M }&\textbf{P1}&\textbf{P2}&\textbf{L1}&\textbf{L2}&\textbf{P1}&\textbf{P2}&\textbf{L1}&\textbf{L2}\\
\hline
\rowcolor{atz_table1}
&\multicolumn{4}{l}{\textbf{5 samples}}&\multicolumn{4}{|l|}{\textbf{3 samples}}\\
\hline
0&0&6&3.28e-01 $\pm$ 4.9e-02 &2.14e-01 $\pm$ 1.1e-02 &0&6&3.67e-01 $\pm$ 5.0e-02 &2.02e-01 $\pm$ 8.7e-03 \\
1&1&7&2.82e-01 $\pm$ 1.3e-02 &2.06e-01 $\pm$ 1.6e-02 &1&7&3.34e-01 $\pm$ 1.9e-02 &2.21e-01 $\pm$ 2.2e-02 \\
2&2&8&3.17e-01 $\pm$ 2.1e-02 &2.60e-01 $\pm$ 8.2e-03 &2&8&3.25e-01 $\pm$ 1.9e-02 &2.17e-01 $\pm$ 2.0e-02 \\
3&3&9&4.55e-01 $\pm$ 5.4e-02 &3.58e-01 $\pm$ 2.9e-02 &3&9&4.44e-01 $\pm$ 3.0e-02 &3.64e-01 $\pm$ 2.7e-02 \\
4&4&10&5.35e-01 $\pm$ 3.3e-02 &4.03e-01 $\pm$ 2.5e-02 &4&10&5.44e-01 $\pm$ 4.4e-02 &3.86e-01 $\pm$ 4.4e-02 \\
\hline
5&5&10&8.29e-01 $\pm$ 7.7e-02 &1.52e-01 $\pm$ 7.0e-02 &5&10&8.73e-01 $\pm$ 8.9e-02 &1.14e-01 $\pm$ 8.2e-02 \\
6&6&7&3.03e-01 $\pm$ 3.4e-02 &1.48e-01 $\pm$ 3.0e-02 &6&7&2.74e-01 $\pm$ 5.4e-02 &1.89e-01 $\pm$ 4.7e-02 \\
7&7&12&2.09e-01 $\pm$ 1.7e-02 &1.85e-01 $\pm$ 4.0e-03 &7&12&2.54e-01 $\pm$ 2.6e-02 &1.87e-01 $\pm$ 9.6e-03 \\
8&8&2&3.55e-01 $\pm$ 2.3e-02 &2.06e-01 $\pm$ 1.4e-02 &8&2&3.38e-01 $\pm$ 2.9e-02 &1.92e-01 $\pm$ 1.0e-02 \\
9&9&14&4.59e-01 $\pm$ 2.3e-02 &2.82e-01 $\pm$ 4.5e-02 &9&14&4.70e-01 $\pm$ 4.4e-02 &2.34e-01 $\pm$ 6.0e-02 \\
\hline
10&10&4&4.83e-01 $\pm$ 3.0e-02 &4.24e-01 $\pm$ 3.3e-02 &\textbf{4} &\textbf{10}&4.30e-01 $\pm$ 4.5e-02 &4.25e-01 $\pm$ 5.4e-02 \\
11&11&12&2.37e-01 $\pm$ 1.1e-02 &1.91e-01 $\pm$ 8.1e-03 &11&12&2.63e-01 $\pm$ 1.2e-02 &1.74e-01 $\pm$ 9.6e-03 \\
12&12&16&2.53e-01 $\pm$ 7.4e-03 &1.72e-01 $\pm$ 7.9e-03 &12&16&2.44e-01 $\pm$ 1.1e-02 &1.52e-01 $\pm$ 8.5e-03 \\
13&13&17&2.91e-01 $\pm$ 2.1e-02 &2.64e-01 $\pm$ 2.8e-02 &\textbf{17} &\textbf{13}&3.17e-01 $\pm$ 5.7e-02 &2.86e-01 $\pm$ 4.2e-02 \\
14&14&9&4.63e-01 $\pm$ 3.5e-02 &2.57e-01 $\pm$ 1.8e-02 &14&9&4.50e-01 $\pm$ 5.1e-02 &2.88e-01 $\pm$ 3.8e-02 \\
\hline
15&15&16&2.31e-01 $\pm$ 3.6e-02 &1.89e-01 $\pm$ 3.5e-02 &15&16&2.15e-01 $\pm$ 3.6e-02 &1.74e-01 $\pm$ 3.1e-02 \\
16&16&19&2.23e-01 $\pm$ 4.8e-03 &1.78e-01 $\pm$ 1.1e-02 &16&19&1.94e-01 $\pm$ 3.5e-03 &1.54e-01 $\pm$ 9.8e-03 \\
17&17&13&3.00e-01 $\pm$ 9.8e-03 &2.66e-01 $\pm$ 1.1e-02 &17&13&3.35e-01 $\pm$ 1.3e-02 &2.96e-01 $\pm$ 1.8e-02 \\
18&18&20&3.35e-01 $\pm$ 1.1e-02 &2.34e-01 $\pm$ 1.9e-02 &18&20&3.41e-01 $\pm$ 1.4e-02 &2.44e-01 $\pm$ 1.6e-02 \\
19&19&16&2.30e-01 $\pm$ 1.9e-02 &2.08e-01 $\pm$ 7.2e-03 &19&16&2.03e-01 $\pm$ 1.7e-02 &2.02e-01 $\pm$ 8.9e-03 \\
20&20&18&3.57e-01 $\pm$ 5.3e-02 &2.87e-01 $\pm$ 5.5e-02 &20&18&3.41e-01 $\pm$ 8.1e-02 &2.63e-01 $\pm$ 8.7e-02 \\
\hline
\end{tabular}
}
\caption{
\textbf{Results for Manifold Shape Identification: Bayesian Inference based on
Laplace-Beltrami Responses.}
We show GNPs can be used for Bayesian estimates of manifold shapes using
observations of the Laplace-Beltrami responses.  We show results for both $5$
and $3$ samples of Laplace-Beltrami pairs $(u, f)$.  The first column \textit{M}
gives the true manifold, and the \textit{P1} and \textit{P2} give the top two
predictions for the manifold.  The columns \textit{L1} and \textit{L2} give the
Bayesian posterior likelihood associated with the predictions of the manifold
shape.  We report the mean $\pm$ the standard deviation of the results over $5$
training runs.  We highlight in bold the cases where the first prediction
disagreed with the true manifold.  We see the case with only $3$ samples had
only two errors.  Interestingly, we see in both sample cases the same two
manifolds appear in the top two predictions, including the correct manifold.
The case with $5$ samples did not have any errors when using the GNPs to
predict the shape.} 
\label{table:shape_id}
\end{center}
\end{table}
 
We test our GNP-Bayesian methods by considering an underlying target true manifold 
$\mathcal{M}^*$ and constructing $M$ observation samples 
for the Laplace-Beltrami responses
$\mathcal{D} = \set{(u^{(i)},
f^{(i)})}_{i = 1}^M$, for $M=3,5$.  For 
each sample $i=1, \dots, M$, we sample the manifold geometry to obtain 
a point-cloud representation $\set{\bsy x_j}_{j = 1}^N$ with $N = 1024$ points.  These
were used to construct observation data $\mathcal{D}$ for testing our 
inference methods. 

We show results of our GNP-Bayesian methods for manifold shapes in 
Table~\ref{table:shape_id}. 
We considered the cases for both $5$ and $3$ samples of the 
Laplace-Beltrami response pairs $(u, f)$.   We report 
in our results the top two predictions \textit{P1}, \textit{P2} for the manifold.  
We also give the  Bayesian posterior likelihood associated with the predictions of the manifold
shape, denoted by \textit{L1} and \textit{L2}.
We report the mean $\pm$ the standard deviation of the results over $5$
training runs. We highlight in bold the cases where the first prediction
disagreed with the true manifold.  We find in the case of using $5$ samples for 
the Laplace-Beltrami responses, we are able to identify 
shapes well.  In this case, there were no errors over 
the $21$ test shapes.  We can see in some cases, such as $10,13$  the 
posterior likelihoods are similar, indicating potentially similar 
Laplace-Beltrami responses.  We find when reducing the number of response
samples to $3$ these two manifolds are misidentified.  Interestingly, we 
do see in both sample cases the same two manifolds appear in the top-two predictions, 
including the correct manifold.  The case with $5$ samples did not have any errors 
when using the GNPs to predict the shape.  These results show that GNPs are 
capable of learning surrogate models for 
Bayesian inference tasks involving significant geometric 
contributions.

\begin{figure}[h]
\centering
\includegraphics[width=0.9\textwidth]{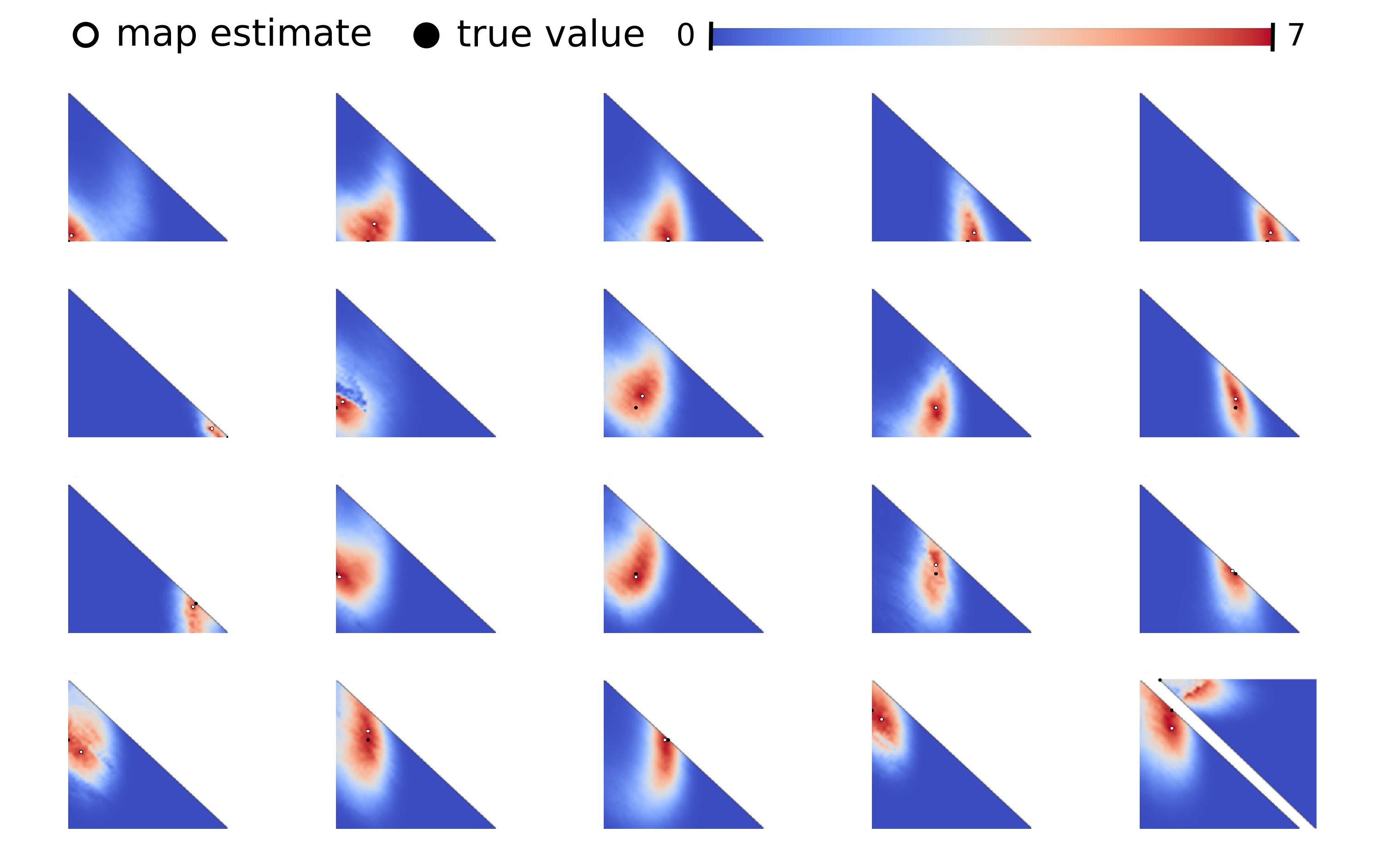}
\caption{\textbf{Manifold Shape Estimation: Solutions of the Bayesian Inverse Problem.}
We estimate the shape of manifolds by solving a Bayesian inverse problem
based on observation of Laplace-Beltrami (LB) responses.  Shown is the 
posterior distribution over the shape space when using for the 
empirical data $5$ samples of the LB-responses.  The results are shown
when the ground-truth shape has index $i$ ranging from $i=0,1,2,\ldots, 20$
with ordering top-to-bottom and left-to-right.
The combined results for the cases $19,20$ are shown on the lower right. }
\label{fig:posterior}
\end{figure}

We also show the full posterior distribution in Figure~\ref{fig:posterior}.  The
full posterior distribution can be used to obtain some further insights into
the predictions.  We find for many shapes there is a wide range in the
posterior distribution for several shapes.  This indicates these likely have
similar responses.  These results reinforce that for some predictions, even if
correctly identified, there may be less overall certainty in the predicted
shape.  This highlights the need for enough sampling and for sufficient
richness of the responses to obtain correct shape identification.  The overall
results show GNPs can be used within Bayesian inverse problems to capture
the roles played by geometry.

\section*{Conclusions}
We have shown how deep learning methods can be leveraged to develop
methods for incorporating geometric contributions into 
data-driven learning of operators.  We showed how geometric 
operators can be learned from manifolds including with 
point-cloud representations.  We showed how geometric operators
can be developed for estimating the metric, curvatures,
differential operators, solution maps to PDEs, and 
shape identification in Bayesian inverse problems.  The developed
Geometric Neural Operators (GNPs) can be used for diverse 
learning tasks where there are significant contributions from 
the geometry.

\newpage
\section*{Acknowledgments}

Authors research supported by grant NSF Grant DMS-2306101.  Authors also would
like to acknowledge computational resources and administrative support at
the UCSB Center for Scientific Computing (CSC) with grants
NSF-CNS-1725797, MRSEC: NSF-DMR-2308708, Pod-GPUs: OAC-1925717, and support from
the California NanoSystems Institute (CNSI) at UCSB.  P.J.A. also would
like to acknowledge a hardware grant from Nvidia.

\newpage
\clearpage

\appendix
\addcontentsline{toc}{section}{Appendices}

\printbibliography

\end{document}